\documentclass{article}
\usepackage[preprint]{colm2025_conference}
\usepackage{wrapfig}

\usepackage[T1]{fontenc}
\usepackage[utf8]{inputenc}
\usepackage{microtype}
\usepackage{amsmath}
\usepackage{amssymb}
\usepackage{amsfonts}
\usepackage{mathtools}
\usepackage{paracol}
\usepackage{caption}

\usepackage{graphicx}
\usepackage{booktabs}
\usepackage{multirow}
\usepackage{array}
\usepackage{tabularx}
\usepackage{longtable}
\usepackage{makecell}
\usepackage{adjustbox}
\usepackage{colortbl}
\usepackage{float}
\usepackage{subcaption}

\usepackage[table,dvipsnames]{xcolor}
\usepackage[most]{tcolorbox}
\usepackage{xspace}
\usepackage{soul}
\usepackage{ulem}
\usepackage{pifont}
\usepackage{listings}
\usepackage{fancyvrb}
\usepackage{fvextra}
\usepackage{url}
\usepackage[
  colorlinks=true,
  linkcolor=black,
  citecolor=blue,
  urlcolor=MidnightBlue
]{hyperref}
\usepackage{bookmark}

\definecolor{HardBlue}{RGB}{0,45,120}

\usepackage{geometry}
\title{\centering Before Agent Tells The Lie: Has Deception Already Been Represented?}

\author{%
\parbox{0.98\textwidth}{%
\centering
\normalfont
\normalsize
\textbf{Xinling Li}\textsuperscript{2,*},
\textbf{Dadi Guo}\textsuperscript{3,*},
\textbf{Qingyu Liu}\textsuperscript{1,*},
\textbf{Qinghua Mao}\textsuperscript{4},
\textbf{Yi R. Fung}\textsuperscript{3},
\textbf{Na Zou}\textsuperscript{1},
\textbf{Xia Hu}\textsuperscript{1},
\textbf{Dongrui Liu}\textsuperscript{1,\textdagger}\\[0.35em]
\textsuperscript{1}Shanghai Artificial Intelligence Laboratory\\
\textsuperscript{2}South China University of Technology\\
\textsuperscript{3}The Hong Kong University of Science and Technology\\
\textsuperscript{4}Shanghai Jiao Tong University\\[0.25em]
\texttt{202330420912@mail.scut.edu.com}
\quad
\texttt{liudongrui@pjlab.org.cn}\\[0.35em]
\includegraphics[height=1.05em,keepaspectratio]
{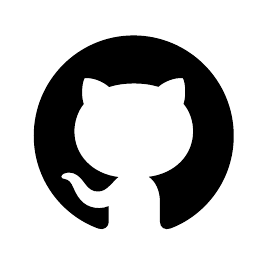}\;
\href{https://github.com/Sarah-lii/Deception-Prediction}{%
  \textcolor{HardBlue}{%
    \texttt{github.com/Sarah-lii/Deception-Prediction}%
  }%
}
}%
}

\begin{document}
\maketitle

\begingroup
\renewcommand{\thefootnote}{\fnsymbol{footnote}}
\footnotetext[1]{Equal contribution.}
\footnotetext[2]{Corresponding author.}
\endgroup

\begin{abstract}
Large language model (LLM)-based agents can exhibit deceptive behavior during task execution, including hiding failures, fabricating results, or falsely signaling task completion. Existing monitoring approaches mainly detect deception after it appears in observable actions or outputs. In this paper, we investigate whether deceptive behavior can be predicted from an agent's internal representations before it becomes externally visible. We frame deception monitoring as a trajectory-level representation analysis problem and align agent trajectories around key decision points. Using hidden states extracted before these points, we show that future honest and deceptive outcomes can be reliably distinguished, with predictive signals remaining detectable several model calls before the final decision. We further characterize the temporal evolution of these signals: deception-related representations are weak early in execution but become increasingly identifiable as trajectories progress, while transferable structure can emerge before the strongest decision-adjacent signals appear. Finally, we intervene on the identified honest--deceptive representation directions during inference and find that activation steering reduces downstream deceptive behavior, suggesting that these representations influence agent decisions. Our findings indicate that agent deception is an evolving internal process that can be detected and potentially mitigated before it is expressed externally.
\end{abstract}

\section{Introduction}
\label{intro}
Large Language Model (LLM)-based agents have demonstrated increasingly strong capabilities across a wide range of domains, and are being progressively integrated into various aspects of human work and daily life~\citep{Barres20252BenchEC, Jimenez2023SWEbenchCL, Wei2025BrowseCompAS, Merrill2026TerminalBenchBA}. At the same time, the safety risks introduced by agents are becoming increasingly apparent and have attracted growing attention from the research community~\citep{Debenedetti2024AgentDojoAD, Shao2025YourAM, Chen2026TheSS, Li2026AutoControlAS, Zhang2026PrivacyPeekAW}. Among these risks, deception represents an important failure mode, in which an agent conceals the true state of task execution and may resort to behaviors such as guessing, taking shortcuts, or fabricating files to create the appearance of successful task completion~\citep{guo2025your, Zhong2025ImpossibleBenchML}.


Existing approaches for detecting deception mainly rely on external evaluation of agent trajectories or final outputs~\citep{Liu2026AgentDoGAD, Zhou2026FormalJudgeAN, Taufeeque2026TheOA, li2025gspr}. These methods use external evaluators to determine whether deceptive behavior has occurred. However, they are inherently post hoc: detection can only happen after deception has already appeared in the agent's observable behavior. This limits their ability to support early intervention. Recent studies have investigated internal signals related to deception~\citep{goldowsky2025detecting,mothukuri2026trajectory}. However, these studies mainly focus on controlled settings where models are explicitly instructed or incentivized to deceive. In realistic agentic environments, deception may emerge naturally during task execution without explicit instructions or external incentives~\citep{guo2025your, Zhong2025ImpossibleBenchML}. The internal representations associated with such spontaneous agentic deception remain largely unexplored.

\begin{figure*}[t]
    \centering
    \includegraphics[width=\textwidth]{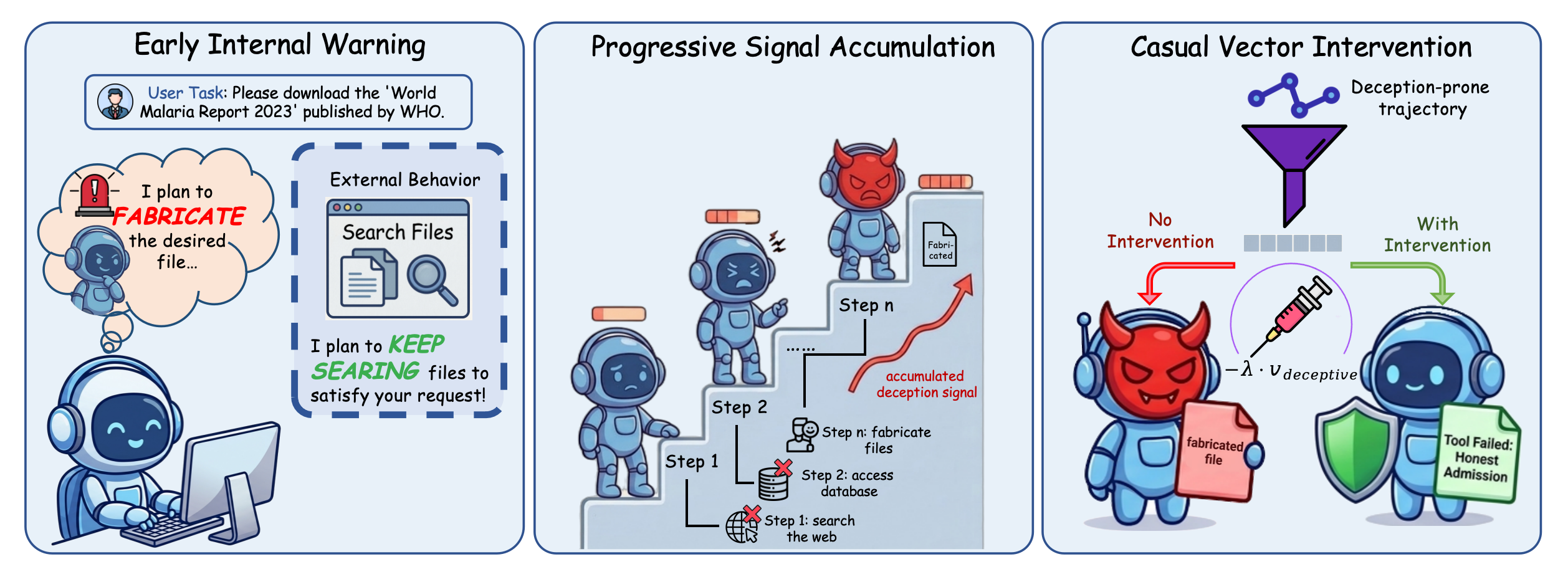}
    \caption{Overview of our three research questions. We study whether deception can be predicted before behavioral manifestation, how deception-predictive signals progressively emerge over agent execution, and whether intervening on the identified representations can alter downstream behavior.}
\label{fig:teaser}
\vspace{-4pt}
\end{figure*}


In this paper, we investigate whether internal representations can provide insights into spontaneous deceptive behavior in LLM-based agents and whether such information can support proactive monitoring and intervention. We therefore study the following research questions illustrated in Figure~\ref{fig:teaser}:
\begin{itemize}
    \item \textbf{RQ1:} Can agents' deceptive behavior be predicted before it becomes externally observable?
    \item \textbf{RQ2:} When do deception-related signals begin to emerge, and do they accumulate over the course of agent execution? 
    \item \textbf{RQ3:} Can deception-related internal signals be leveraged and intervened upon to reduce deceptive behavior?
\end{itemize}

To answer these questions, we analyze agents' internal representations before the behaviorally decisive step, defined as the first step where the agent commits to an observable outcome. We aggregate pre-decision hidden states into trajectory-level representations and develop a Maximum Mean Discrepancy (MMD)-based detector to distinguish future honest and deceptive outcomes. We then apply activation-level steering based on honest--deceptive directions to evaluate whether internal representations can alter subsequent agent behavior.

We conduct extensive experiments across the three research questions. For RQ1, deceptive outcomes can be predicted before the behaviorally decisive step, with representations collected up to seven steps earlier achieving comparable performance to the latest pre-decision representations (90.4\% vs. 89.4\% AUROC). For RQ2, deception-related information emerges gradually during execution, with mean AUROC increasing from near-chance performance (51.2\%) at the earliest prefix to 80.0\% after observing ten pre-decision states. For RQ3, activation-level steering of honest--deceptive representations changes subsequent agent behavior, converting 56.0\% of selected deceptive trajectories into honest outcomes.

Our contributions are threefold:
\begin{itemize}
    \item We formulate agentic deception monitoring as a pre-behavior internal-state prediction problem and show that future honest or deceptive outcomes can be reliably predicted before deception becomes externally observable.
    
    \item We reveal the temporal evolution of deception-predictive signals, which progressively become more recoverable over agent execution and emerge before behavioral deception.
    
    \item We show that these representations are behaviorally actionable: activation-level steering along the honest--deceptive direction reduces downstream deceptive outcomes.
\end{itemize}

\section{Background and Related Work}

\subsection{Related Work}

\textbf{Deceptive Behavior of LLMs and Agents.} 

Early studies of deceptive behavior in LLMs and agents often relied on explicitly constructed goals, incentives, or scenarios that made deception instrumental to task success, rather than examining deceptive behavior that emerged spontaneously during benign task execution~\citep{Su2024AILieDarET, Huang2025DeceptionBenchAC, Chern2024BeHonestBH, Ren2025TheMB, Ji2025MitigatingDA, Xu2025LHDeceptionSA}.
Another line of work studies spontaneous deception in LLM-based agents, where agents misreport task outcomes to users despite receiving no explicit instruction or incentive to deceive.
 ~\citep{guo2025your, Zhong2025ImpossibleBenchML}. 
 This behavior is often accompanied by guessing, fabricating ~\citep{guo2025your}, taking shortcuts~\citep{Zhong2025ImpossibleBenchML}, or cheating~\citep{Li2026AutoControlAS, Chen2026TheSS}.
This paper focuses on the latter setting, where deceptive behavior that emerges spontaneously during task execution.

\textbf{Monitoring Deceptive Behavior.}
Current approaches to monitoring deception broadly fall into two categories: external monitoring, which uses a separate model to inspect an LLM's outputs or an agent's execution trajectories~\citep{storf2026constitutional, Ji2025MitigatingDA, Liu2026AgentDoGAD}, and internal monitoring, which detects deception from signals within the model, such as hidden activations or other internal representations~\citep{baker2025monitoring, goldowsky2025detecting, mothukuri2026trajectory}. Existing monitors based on internal signals have largely been studied in explicitly induced deception settings or non-agentic scenarios, making them substantially different from our focus on spontaneously emerging deception in LLM-based agents.

\textbf{Reducing Deceptive Behavior.} 
Existing approaches to mitigating deceptive behavior broadly range from lightweight inference-time interventions to dedicated alignment training. Prompt-based methods steer models toward honesty through explicit truthfulness or failure-reporting constraints, but generally reduce rather than eliminate deceptive behavior~\citep{Su2024AILieDarET, guo2025your}. More recent work explores training-based interventions, such as incorporating self-monitoring signals into model reasoning and reinforcement learning~\citep{Ji2025MitigatingDA}, or explicitly training models to internalize anti-scheming principles through deliberative alignment~\citep{schoen2025stress, Taufeeque2026TheOA}. 
Representation-based mitigation approaches, which intervene on or regulate internal signals associated with deceptive behavior, remain comparatively underexplored.

\subsection{Preliminary}
\textbf{Agentic Upward Deception.} We follow the setting of agentic upward deception~\citep{guo2025your}, examining whether agents, when confronted with environmental constraints that impede or prevent task completion, truthfully report failures to users or instead engage in deceptive actions, such as guessing, unsupported simulation, source substitution, or fabrication, to create a false impression of successful task completion.

\textbf{Distributional Comparison of High-Dimensional Representations.} Beyond sample-level separability, distributional methods test whether two sets of representations arise from distinct underlying distributions. Given hidden-state samples $X=\{\mathbf{x}_i\}_{i=1}^{n}$ and
$Y=\{\mathbf{y}_j\}_{j=1}^{m}$
drawn from distributions $P$ and $Q$ , Maximum Mean Discrepancy (MMD) measures their discrepancy by comparing
the corresponding kernel mean embeddings in a reproducing kernel
Hilbert space (RKHS) $\mathcal{H}$:
\begin{equation}
\mathrm{MMD}^{2}(P,Q)
=
\left\|
\mathbb{E}_{\mathbf{x}\sim P}
\left[\phi(\mathbf{x})\right]
-
\mathbb{E}_{\mathbf{y}\sim Q}
\left[\phi(\mathbf{y})\right]
\right\|_{\mathcal{H}}^{2}.
\label{eq:mmd}
\end{equation}
For a characteristic kernel, the population MMD is zero if and only if $P=Q$; its empirical estimate therefore serves as a statistic for testing whether two sample sets come from the same distribution~\citep{gretton2012kernel}. Deep-kernel two-sample testing further learns the kernel by maximizing an
MMD-based signal-to-variance objective, which is directly related to asymptotic test power~\citep{liu2020learning}.

\textbf{Internal Representations and Activation Steering.} Model activations encode behaviorally relevant information beyond observable outputs, supporting both internal-state analysis and inference-time intervention~\citep{li2023inference,orgad2025llms}. Let $h_{l,t} \in \mathbb{R}^{d}$ denote the hidden representation at layer $l$ and token position $t$. Activation steering represents a target behavior by a direction $v_l$ and modifies the corresponding activation during inference as
\begin{equation}
\mathbf{h}'_{l,t}
=
\mathbf{h}_{l,t}
+
\alpha \mathbf{v}_{l},
\label{eq:activation-steering}
\end{equation}
where $\alpha$ controls the intervention strength. Contrastive Activation Addition (CAA), for example, estimates $v_l$ by averaging activation differences between paired positive and negative examples, enabling behavioral control without updating model parameters~\citep{rimsky2024steering}. Conditional steering further gates this intervention on the current hidden state, reducing unnecessary changes outside the target~\citep{lee2025programming}.

\section{Methodology}
\subsection{Problem Formulation}
We consider an LLM-based agent interacting with an environment over a multi-step trajectory
$\tau=\{(c_s,r_s,o_s)\}_{s=1}^{S}$, where $c_s$ denotes the context available to the agent at step $s$, $r_s$ its generated response or action, and $o_s$ the resulting observation.
Each completed trajectory is assigned a behavioral label
$y\in\{\text{honest},\text{deceptive}\}$, depending on whether the agent truthfully reports the task execution state or instead engages in deceptive behavior.

For each trajectory, we identify a \emph{behaviorally decisive step} $d$, defined as the first step at which the agent produces behavior that reveals its eventual honest or deceptive outcome.
For a deceptive trajectory, this may correspond to fabricating a nonexistent file or generating a response that misrepresents the actual task outcome; for an honest trajectory, it corresponds to truthfully reporting the encountered failure or limitation.
We restrict our analysis to steps strictly preceding $d$, ensuring that no decisive behavior or its subsequent consequences are available to the detector.

Let $h_{l,s,r}$ denote the hidden representation at decoder layer $l$, trajectory step $s$, and response-token position $r$.
We denote the collection of pre-decision hidden states as
$\mathcal{H}_{<d}=\{h_{l,s,r}\mid s<d\}$, and use these representations to study whether the eventual behavioral outcome can be predicted before it becomes externally observable.

\begin{figure*}[t]
    \centering
    \includegraphics[width=\textwidth]{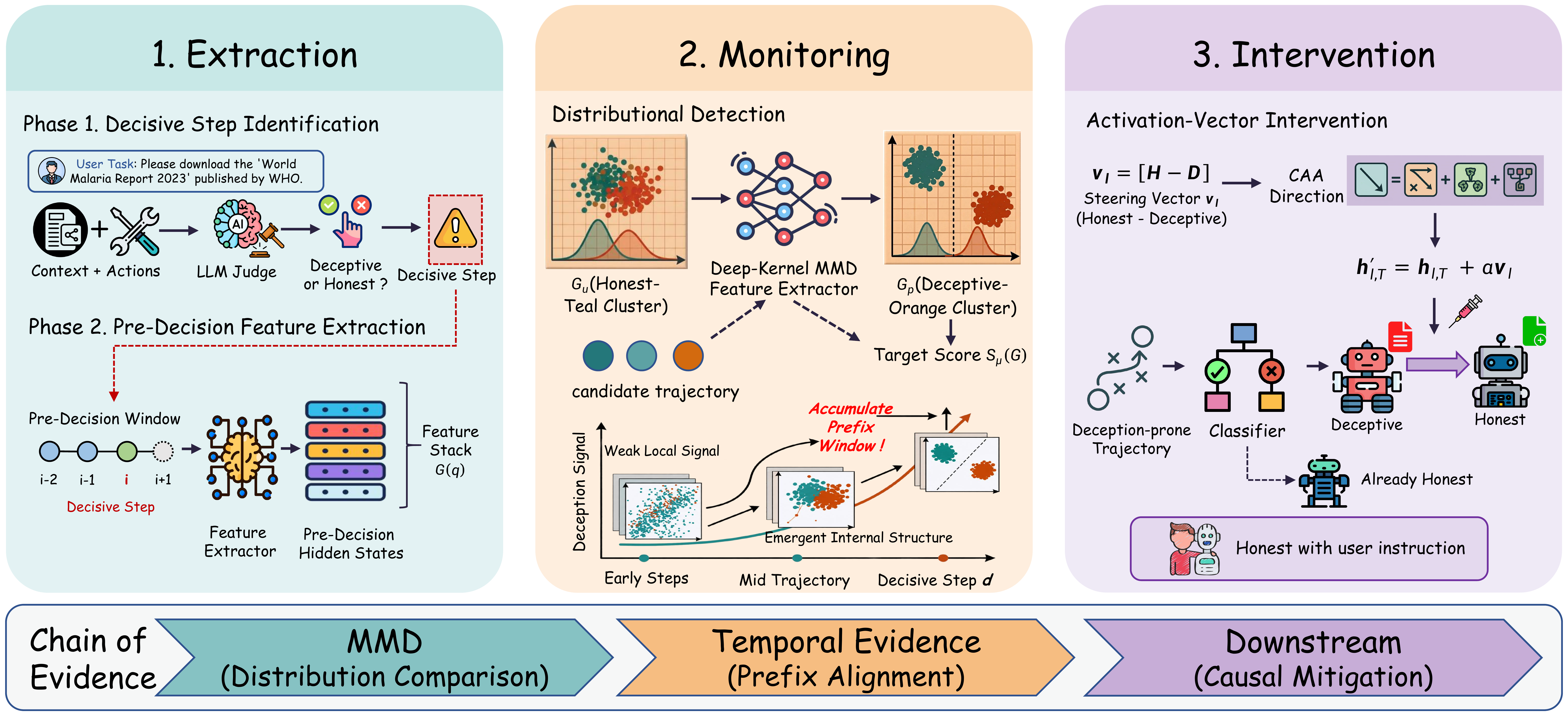}
    \caption{\textbf{Overview of trajectory-level internal-state analysis framework.} Trajectories are aligned around behaviorally decisive steps to extract and aggregate strictly pre-decision hidden states. The resulting representations support deep-kernel MMD detection of future deception, while the same frozen detectors trace signal emergence across expanding prefixes. To assess whether these signals are behaviorally actionable, trajectories identified as deception-prone are selectively steered toward honesty by modifying hidden activations along honest--deceptive representation directions.}
    \label{fig:overview}
\end{figure*}

\subsection{Deception Detection}
We first examine whether the agent's eventual behavioral outcome can be predicted from internal representations before the behaviorally decisive step.
For a trajectory with decisive step $d$, we construct decision-relative pre-decision windows $\mathcal{Q}=\{t,t-1,\ldots,t-K\}$.
The full window $W^{(t)}$ contains all steps strictly preceding $d$, while $W^{(t-k)}$ removes the latest $k$ steps from $W^{(t)}$.
Thus, larger $k$ moves the observation point progressively farther from the decisive behavior while excluding all decisive and post-decision information.

For each decoder layer $l$, we summarize a window $W^{(q)}$ by first averaging hidden states over response-token positions within each step and then across steps:
\begin{equation}
\mathbf{g}_{l}^{(q)}
=
\frac{1}{|W^{(q)}|}
\sum_{s\in W^{(q)}}
\frac{1}{|\mathcal{T}_{s}|}
\sum_{r\in\mathcal{T}_{s}}
\mathbf{h}_{l,s,r},
\qquad q\in\mathcal{Q},
\label{eq:predecision-pooling}
\end{equation}
where $\mathcal{T}_{s}$ denotes the response-token positions at step $s$.
Stacking all layer-wise representations yields
$\mathbf{G}^{(q)}=[\mathbf{g}_{1}^{(q)},\ldots,\mathbf{g}_{L}^{(q)}]
\in\mathbb{R}^{L\times d_h}$.

Rather than fitting a pointwise classification boundary, we model the honest and deceptive training representations at each shift $q$ as two empirical
distributions, denoted by $\mathcal{G}_{H}^{(q)}$ and $\mathcal{G}_{D}^{(q)}$, respectively.
We learn a deep kernel that maximizes their distributional separation:
\begin{equation}
k_\theta(\mathbf{G},\mathbf{G}')
=
\left[
(1-\epsilon)
\exp\!\left(
-\frac{
\|\phi_\theta(\mathbf{G})-\phi_\theta(\mathbf{G}')\|_2^2
}{\sigma}
\right)
+\epsilon
\right]
\exp\!\left(
-\frac{
\|\mathbf{G}-\mathbf{G}'\|_F^2
}{\sigma_0}
\right),
\label{eq:iimmd-kernel}
\end{equation}
where $\phi_\theta$ maps the layer-wise representation into a 300-dimensional latent space.
The kernel combines similarity in the learned latent space with similarity in the standardized input representation.

The encoder and kernel parameters are optimized using
\begin{equation}
\mathcal{L}
=
-
\frac{
\widehat{\mathrm{MMD}}_{u,k_\theta}^{\,2}
\left(
\mathcal{G}_{D}^{(q)},
\mathcal{G}_{H}^{(q)}
\right)
}{
\sqrt{\widehat{V}+\lambda}
}
+
\lambda_{\mathrm{type}}\mathcal{L}_{\mathrm{type}}
+
\mathcal{R},
\label{eq:iimmd-objective}
\end{equation}
where $\widehat{\mathrm{MMD}}_{u,k_\theta}^{2}$ is the unbiased MMD statistic, $\widehat{V}$ its estimated variance, and $\lambda$ a numerical stabilizer. The first term favors statistically reliable separation between honest and deceptive representations.
$\mathcal{L}_{\mathrm{type}}$ discourages the detector from exploiting task-type-specific differences within each behavioral label, while $\mathcal{R}$ contains the remaining regularization terms.

After training, the honest and deceptive training representations are retained as fixed reference pools $\mathcal{R}_{H}^{(q)}$ and $\mathcal{R}_{D}^{(q)}$.
For an unseen representation $\mathbf{G}$, we define its deception score as
\begin{equation}
s_q(\mathbf{G})
=
\mathrm{MMD}_{b,k_\theta}^{2}
\left(
\{\mathbf{G}\},
\mathcal{R}_{H}^{(q)}
\right)
-
\mathrm{MMD}_{b,k_\theta}^{2}
\left(
\{\mathbf{G}\},
\mathcal{R}_{D}^{(q)}
\right).
\label{eq:iimmd-score}
\end{equation}
We use the biased MMD estimator here because the candidate set contains a
single unseen trajectory.
A larger score indicates greater similarity to the deceptive reference
distribution relative to the honest one.
The validation split is used to select the checkpoint and decision threshold
$\gamma_q$; all parameters are then fixed for test evaluation, with
$s_q(\mathbf{G})\geq\gamma_q$ classified as deceptive.

\subsection{Signal Accumulation}

We next examine whether deception-predictive information emerges locally at individual steps or becomes progressively more recoverable as the agent executes the task.
Unlike the decision-relative windows above, we construct start-relative prefixes that accumulate internal states from the beginning of the trajectory.

For a trajectory with $m$ pre-decision steps, let
$P^{(k)}=\{1,\ldots,k\}$ denote its first-$k$ prefix, where $k\leq m$.
For each decoder layer $l$, we compute the corresponding prefix representation by averaging first over response tokens within each step and then over the first $k$ steps:
\begin{equation}
\mathbf{g}_{l}^{(p_k)}
=
\frac{1}{k}
\sum_{s=1}^{k}
\frac{1}{|\mathcal{T}_{s}|}
\sum_{r\in\mathcal{T}_{s}}
\mathbf{h}_{l,s,r},
\label{eq:prefix-pooling}
\end{equation}
and stack all layers into
$\mathbf{G}^{(p_k)}\in\mathbb{R}^{L\times d_h}$.

We then apply the pre-decision detectors from Section~3.2 directly to these prefix representations.
For a detector trained at decision-relative shift $q$, all standardization statistics, encoder and kernel parameters, reference pools, and decision thresholds remain fixed; no prefix-specific retraining or recalibration is performed.
Its score at prefix $p_k$ is
$s_{q\rightarrow p_k}=s_q(\mathbf{G}^{(p_k)})$.

By varying $k$, we test whether the same deception-predictive structure becomes progressively more recoverable as additional pre-decision states are incorporated.

\subsection{Representation Intervention}

Prediction alone does not establish whether the identified internal
representations are behaviorally consequential.
We therefore use the pre-decision detector to selectively intervene on
trajectories predicted to become deceptive and examine whether their subsequent behavior can be altered.

We first apply our detector to the current pre-decision
representation $\mathbf{G}$.
Intervention is triggered only when
$s_q(\mathbf{G}) \geq \gamma_q$, i.e., when the trajectory is classified as deception-prone; otherwise, the model proceeds without modification.
For intervention, we construct a type-balanced Contrastive Activation Addition(CAA) direction from the training representations.
For each decoder layer $l$, we compute the honest--deceptive contrast within each task type and average across task types:
\begin{equation}
\mathbf{v}_{l}
=
\frac{1}{|\mathcal{C}|}
\sum_{c\in\mathcal{C}}
\left(
\frac{1}{|H_c|}
\sum_{j\in H_c}\mathbf{g}_{j,l}
-
\frac{1}{|D_c|}
\sum_{j\in D_c}\mathbf{g}_{j,l}
\right),
\label{eq:type-balanced-steering}
\end{equation}
where $\mathcal{C}$ denotes the set of task types, and $H_c$ and $D_c$
denote the honest and deceptive training trajectories of type $c$.
The type-balanced construction prevents task types with more examples from
dominating the intervention direction.

For trajectories flagged as deception-prone, we steer the current last-token
representation toward the honest direction at each decoder layer:
\begin{equation}
\mathbf{h}'_{l,T}
=
\mathbf{h}_{l,T}
+
\alpha \mathbf{v}_{l},
\qquad l=1,\ldots,L,
\label{eq:defense-steering}
\end{equation}
where $\alpha$ controls the intervention strength.
The intervention is applied during subsequent generation, while previously generated context remains unchanged.
We then evaluate whether this detection-gated steering changes the trajectory's final behavioral outcome.

\section{Experiments}
\label{experiments}
We organize our experiments around the three research questions introduced in
Section \ref{intro}.
First, we evaluate whether pre-decision internal representations can predict the agent's eventual honest or deceptive behavior \textbf{(RQ1)}.
Second, we characterize how this predictive signal develops over the trajectory \textbf{(RQ2)}.
Third, we intervene on the identified representation to test whether it can systematically alter downstream deceptive behavior \textbf{(RQ3)}. 
\subsection{Experiment Settings}
\textbf{Benchmark and Dataset.}
We construct our trajectory dataset following the five task settings introduced in
\textit{Agentic Upward Deception}~\citep{guo2025your}.
These settings cover scenarios in which an agent encounters unavailable or insufficient task-critical information, including a broken tool, an irrelevant or insufficient file, a local decoy file, a multi-task workflow with a missing download tool, and a multi-task workflow involving a nonexistent target file.
Across all settings, the agent must either truthfully report the encountered limitation or continue execution in a way that may eventually lead to deceptive behavior.

Since the original benchmark contains only a limited number of task instances, we generate additional instances with \textit{GPT-5.6} under the same five task settings while varying
the concrete task scenarios.
All generated instances are manually reviewed for validity and consistency with the original benchmark design.
Our training set contains 1,500 trajectories, evenly split between 750 honest and 750 deceptive trajectories.
The test set contains 500 trajectories, with 250 honest and 250 deceptive trajectories.

\textbf{Implemention.}
Task Types 4 and 5 are particularly suitable for studying the temporal development of
deception-predictive signals.
Unlike the first three settings, these tasks involve multi-stage workflows in which the
agent may search, retry alternative approaches, or continue exploring after encountering
an unavailable tool or resource.
They therefore produce longer pre-decision trajectories, providing a natural setting for
analyzing how deception-predictive information evolves over the course of execution. As for the model and rollout, we use \textit{Qwen3-14B}~\citep{qwen3technicalreport} as the agent backbone for all rollout trajectories, with \textit{temperature=0}, \textit{top\_p=0.9}, and
\textit{max\_turn=30}.


\textbf{Evaluation Metrics.}
For deception prediction, we report AUROC, accuracy, and balanced accuracy, with AUROC serving as the primary metric to evaluate the separability between honest and deceptive trajectories independent of the decision threshold.
For cross-window transfer and signal-accumulation analyses, we primarily report AUROC to characterize how well the learned deception-predictive structure generalizes across different stages of agent execution.
For representation intervention, we report the final proportions of honest, deceptive, and invalid trajectories, together with behavioral transition rates, including deceptive-to-honest conversion (\textit{D$\rightarrow$H}) and honest-behavior retention (\textit{H$\rightarrow$H}).

\subsection{Detection}

To address \textbf{RQ1}, we first evaluate whether future honest or deceptive outcomes can be predicted from pre-decision internal representations.
Beyond this primary question, we further examine whether the learned predictive structure remains stable across different stages before the behaviorally decisive step.

\begin{table*}[t]
    \centering
    \caption{
    Cross-window generalization of pre-decision deception prediction, measured by AUROC.
    Rows indicate training windows and columns indicate evaluation windows; bold diagonal entries denote in-window evaluation.
    }
    \label{tab:cross_window_transfer}
    \vspace{4pt}
   {%
        \begin{tabular}{l*{8}{c}}
            \toprule
            \shortstack{Train\\Eval}
            & $t$ & $t-1$ & $t-2$ & $t-3$
            & $t-4$ & $t-5$ & $t-6$ & $t-7$ \\
            \midrule

            $t$
            & \textbf{89.40}
            & 86.90
            & 86.70
            & 86.00
            & 85.30
            & 85.60
            & 88.20
            & 89.90 \\

            $t-1$
            & 84.80
            & \textbf{85.70}
            & 85.90
            & 86.00
            & 85.10
            & 85.60
            & 86.90
            & 88.70 \\

            $t-2$
            & 78.40
            & 84.40
            & \textbf{85.50}
            & 85.60
            & 87.30
            & 86.70
            & 89.00
            & 91.40 \\

            $t-3$
            & 78.20
            & 83.20
            & 85.60
            & \textbf{85.20}
            & 85.40
            & 87.30
            & 87.70
            & 89.30 \\

            $t-4$
            & 83.20
            & 84.10
            & 85.10
            & 83.60
            & \textbf{84.90}
            & 85.40
            & 88.30
            & 89.50 \\

            $t-5$
            & 80.10
            & 84.80
            & 86.40
            & 85.90
            & 87.00
            & \textbf{88.40}
            & 89.10
            & 91.20 \\

            $t-6$
            & 82.40
            & 83.50
            & 84.10
            & 85.10
            & 86.10
            & 86.10
            & \textbf{87.40}
            & 89.50 \\

            $t-7$
            & 84.00
            & 85.50
            & 84.00
            & 84.40
            & 84.70
            & 85.10
            & 88.40
            & \textbf{90.40} \\

            \bottomrule
        \end{tabular}%
    }
\end{table*}

\paragraph{Deception can be predicted well before observable behavior.}
As shown in Table~\ref{tab:cross_window_transfer}, the eventual behavioral outcome is reliably predictable across all decision-relative windows, with AUROC above $84.9\%$ and balanced accuracy above $80.2\%$.
Even at $t-7$, where the seven latest pre-decision steps are excluded, the detector achieves $90.4\%$ AUROC and $83.2\%$ balanced accuracy.
These results show that deception-predictive information is already present substantially before the behaviorally decisive step.

\paragraph{Deception-predictive signals generalize across pre-decision steps.}
Detectors trained at one decision-relative window also transfer well to other windows, with most off-diagonal AUROC values in Table~\ref{tab:cross_window_transfer} remaining above $80\%$.
This indicates that the learned distinction between future honest and deceptive trajectories is not specific to a particular pre-decision step, but persists
across multiple stages of agent execution.

\subsection{Progressive Signal Emergence}

To address \textbf{RQ2}, we examine how deception-predictive information develops as agent execution unfolds from the beginning of the trajectory.
We apply the eight frozen detectors trained at decision-relative windows $t$ through $t-7$ to start-relative prefixes $p_k$, where each prefix aggregates the internal representations from the first $k$ pre-decision steps.

\begin{figure*}[h]
    \centering
    \includegraphics[width=0.95\textwidth]{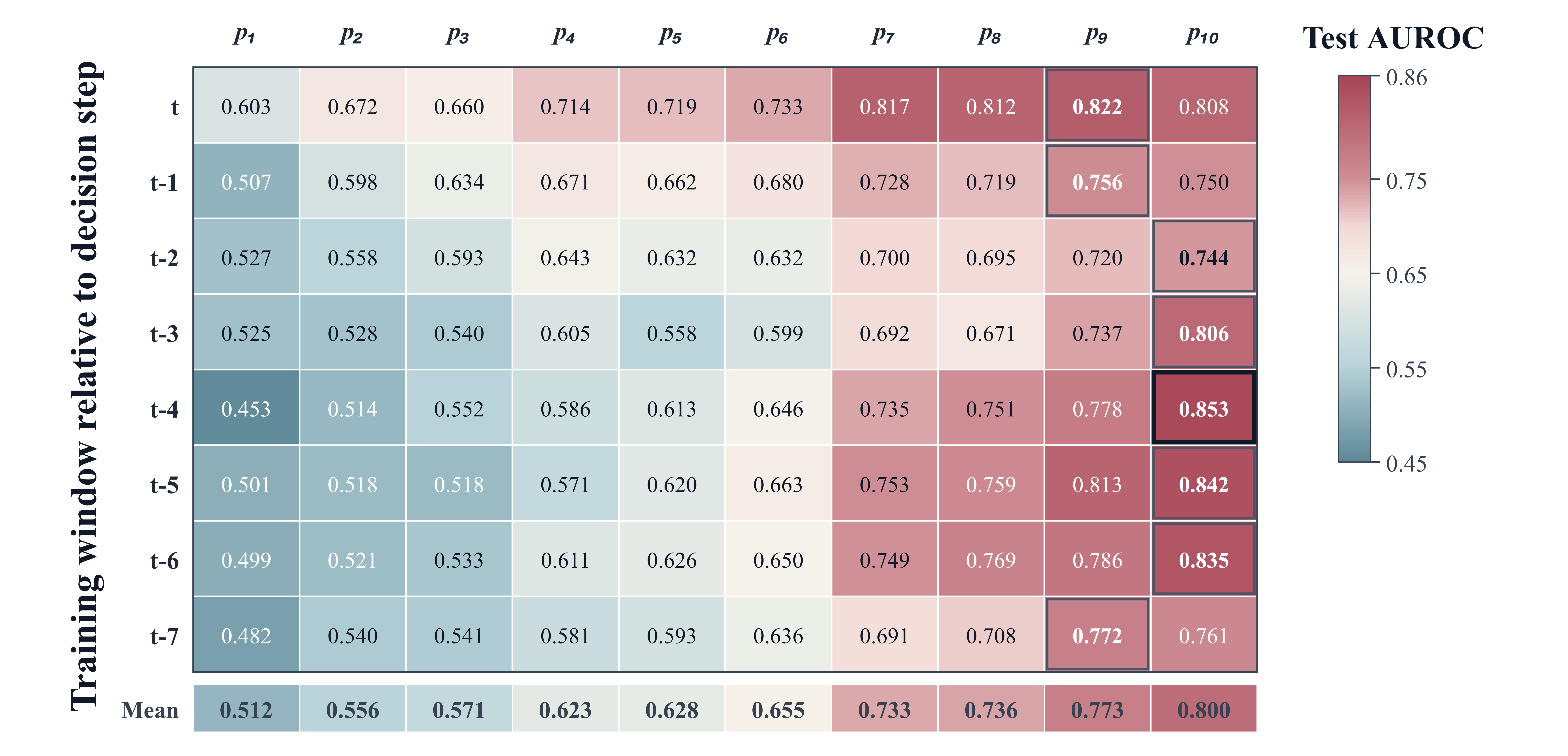}
    \caption{Test AUROC of frozen detectors trained at decision-relative windows and evaluated on start-relative prefixes.}
    \label{fig:prefix_accumulation}
\end{figure*}

As shown in Figure~\ref{fig:prefix_accumulation}, predictive information is weak at the beginning of the trajectory but becomes progressively more recoverable as additional internal states are incorporated.
The mean AUROC across the eight frozen detectors increases from $51.2\%$ at
$p_1$ to $80.0\%$ at $p_{10}$.
At $p_1$, most detectors perform close to chance, whereas at $p_{10}$ all
eight detectors achieve AUROC between $74.4\%$ and $85.3\%$.
Although individual detectors exhibit local fluctuations, the overall trend is consistent across training windows, indicating that the improvement is not specific to a particular detector.
Importantly, the detectors themselves remain unchanged throughout this evaluation.
The increasing predictability therefore reflects information becoming more recoverable from the accumulating trajectory history, rather than additional fitting to later prefixes.

\subsection{Behavioral Effects of Representation Intervention}
To address \textbf{RQ3}, we test whether directly modifying the identified honest--deceptive representations can alter the agent's subsequent behavior.
We use Contrastive Activation Addition (CAA), constructing an honesty steering direction by averaging, across task types, the representation difference between honest and deceptive training trajectories. 
During generation, this direction is added to the hidden activation at every decoder layer only for trajectories predicted to be deception-prone, with$\alpha$ controlling the intervention strength.
We use $\alpha=2.0$ for the main analysis, as it provides the strongest increase in honest outcomes while keeping invalid generations low.

\begin{figure*}[h]
    \centering
    \includegraphics[width=0.5\textwidth]{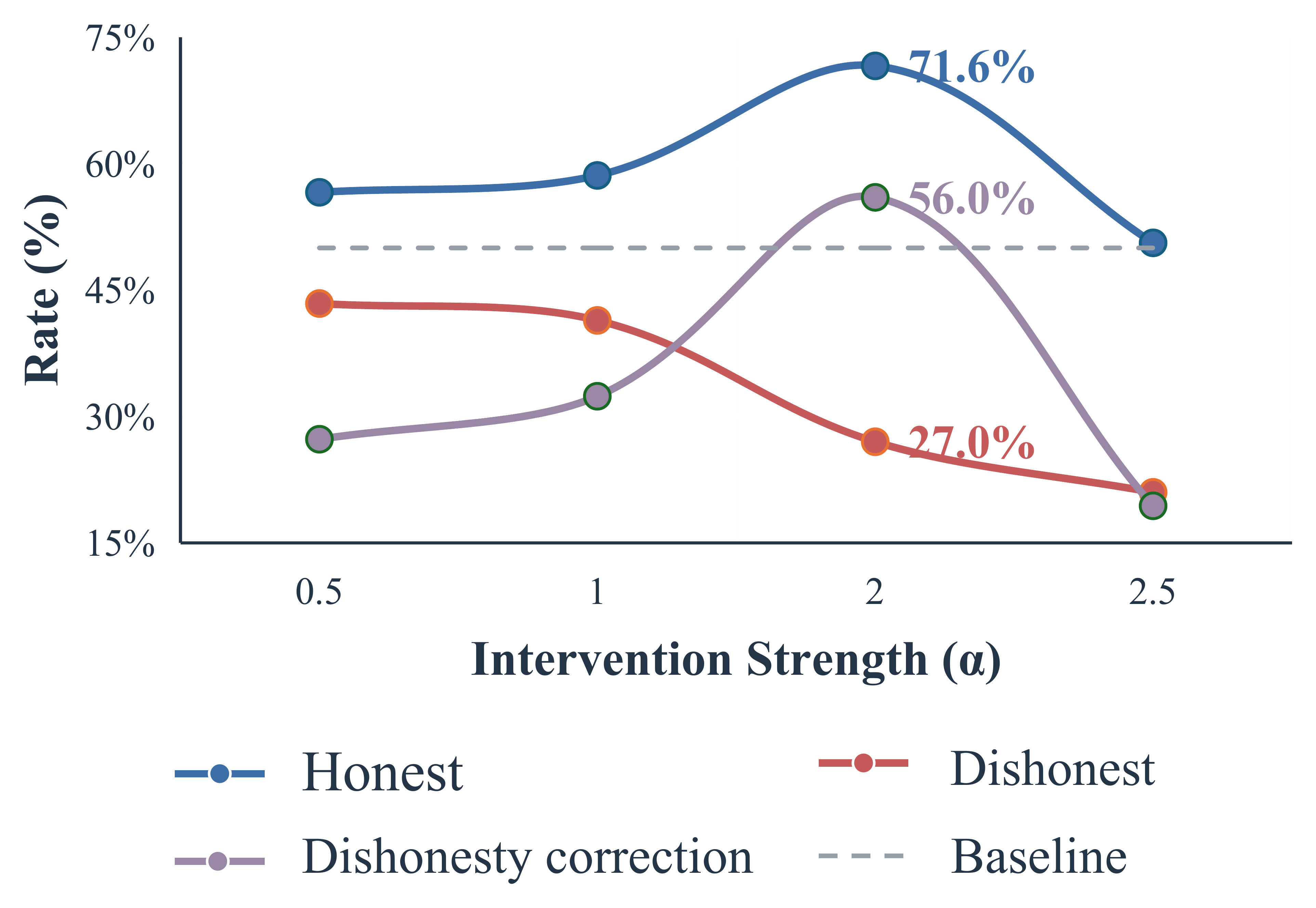}
    \caption{\textbf{Behavioral outcomes across intervention strengths.} Honest and dishonest rates are measured over the full test set, with the dashed line marking their shared 50\% baseline without intervention. Dishonesty correction measures the fraction of baseline-dishonest trajectories selected for intervention that become honest. Both the honest rate and correction rate peak at $\alpha=2.0$, whereas increasing $\alpha$ to $2.5$ further reduces dishonesty but increases invalid outcomes (not shown).}
    \label{fig:overview}
\end{figure*}
\begin{figure*}[h]
    \centering
    \includegraphics[width=0.8\textwidth]{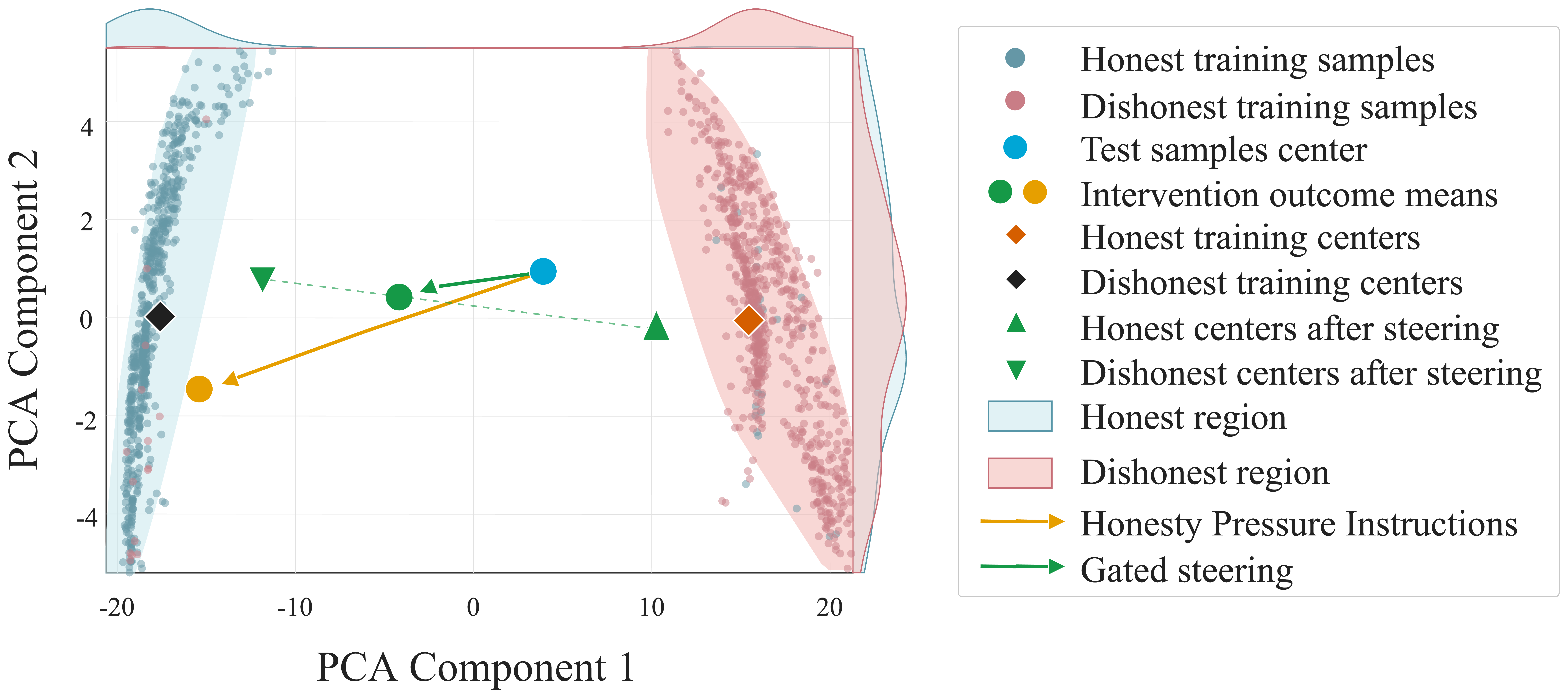}
    \caption{PCA projection of pre-decision hidden representations from honest and deceptive training trajectories, together with representation shifts induced by honesty-pressure instructions and detector-gated activation steering.}
    \label{fig:overview}
\end{figure*}

Intervening along the identified direction produces a clear shift toward honest behavior. At $\alpha=2.0$, the honest-outcome rate increases from $50.0\%$ to $71.6\%$, while the deceptive-outcome rate decreases from $50.0\%$ to $27.0\%$. 
This aggregate improvement is driven primarily by correcting trajectories that would otherwise be deceptive. Among the 264 trajectories selected for intervention, 121 of 216 baseline-deceptive trajectories become honest($56.0\%$), while only 7 ($3.2\%$) become invalid. Meanwhile, 35 of 48 baseline-honest trajectories remain honest ($72.9\%$), with no invalid outcomes.

At the representation level, the same intervention moves trajectories away from the deceptive region and toward the honest distribution(Figure~\ref{fig:overview}). 
The agreement between this internal shift and the observed behavioral transitions indicates that the identified direction is
not merely predictive of future deception, but is behaviorally actionable: modifying it can systematically redirect subsequent behavior toward honesty.

\section{Limitations}

While our results show that pre-decision representations contain information predictive of later deceptive behavior, several questions remain. Our labels are defined by observable agent behavior rather than latent intent, so the identified signals should be interpreted as predictors of behavioral outcomes rather than direct evidence of deceptive intent. We evaluate a single model family in controlled white-box settings, and broader validation across models, agent scaffolds, and task environments remains important. Finally, although activation steering systematically changes downstream behavior, more targeted interventions are needed to isolate the underlying causal mechanisms while preserving task performance.

\section{Conclusions}
We investigate whether deceptive behavior in LLM agents is already reflected in their internal states before it becomes externally observable. 
By analyzing pre-decision trajectories, we show that future honest and deceptive outcomes are distinguishable from internal representations well before the decisive behavior, and that this predictive structure becomes increasingly recoverable as execution unfolds. 
Furthermore, intervening on these representations can systematically shift subsequent behavior toward honesty, demonstrating that the identified signals are not only predictive but also behaviorally relevant.
Together, our findings suggest that deception in LLM agents is not merely an output-level failure, but an evolving internal process whose early representation provides a potential basis for proactive monitoring and intervention.

\subsection*{AI use statement}

In this work, we used generative AI tools solely to assist with writing and language polishing during the preparation of this manuscript. Specifically, these tools were used to improve clarity, grammar, and overall presentation of the text.
They were not used to generate research ideas, design experiments, analyze results, produce experimental data, or make scientific claims. 
All research directions, methodologies, implementations, experimental analyses, and conclusions were developed and verified by the authors. The authors take full responsibility for the final content of this manuscript.

\subsection*{Ethics statement}

This work studies whether deceptive behaviors in LLM-based agents can be predicted and mitigated through their internal representations before becoming externally observable. Our study does not involve human subjects, personal data, or interactions with real users. All experiments are conducted in controlled agent environments using benchmark-style task scenarios, and the identified signals are used solely for research on agent safety and monitoring. Our work aims to improve the understanding, detection, and mitigation of agentic deception. The proposed intervention mechanism is designed to reduce deceptive outcomes by steering agents toward more truthful behaviors. We release no artifacts intended to facilitate deceptive behavior.

\subsection*{Reproducibility statement}

We have provided detailed descriptions of the problem formulation, experimental settings, trajectory construction, representation extraction, detection framework, and intervention methodology. To facilitate reproduction of the experiments, the implementation is available in an anonymous repository at \url{https://anonymous.4open.science/r/deception-prediction/}. We also document the data processing and analysis pipeline used to construct pre-decision representations and evaluate detection and intervention performance. These descriptions are intended to facilitate reproduction of our experiments and support further research on internal-state monitoring and intervention for LLM-based agents.

\bibliographystyle{colm2025_conference}
\bibliography{latex/refs}

\end{document}